%% file: emnlp2018.tex
\documentclass[11pt,a4paper]{article}
\usepackage[hyperref]{emnlp2018}
\usepackage{times}
\usepackage{latexsym}
\usepackage{afterpage}
\usepackage{subfigure}
\usepackage{mathrsfs}
\usepackage{amsmath}
\usepackage{amsfonts}
\usepackage{multirow}
\usepackage{soul}
\usepackage{lipsum}
\usepackage{float}
\usepackage{url}
\usepackage{tikz}
\usetikzlibrary{bayesnet}
\usetikzlibrary{patterns}
\usepackage{slashbox}
\newcommand{\bc}[1]{}
\DeclareMathOperator*{\argmax}{arg\,max}
\DeclareMathOperator*{\argmin}{arg\,min}

\aclfinalcopy

\title{QuaSE: Sequence Editing under Quantifiable Guidance\thanks{~~The work described in this paper was done when Yi Liao was an intern at Tencent AI Lab. The work is partially supported by a grant from the Research Grant Council of the 
Hong Kong Special Administrative Region, China (Project Code: CUHK413510)}}
\author{
Yi Liao$^{13}$, Lidong Bing$^2$, Piji Li$^{12}$, Shuming Shi$^2$, Wai Lam$^1$, Tong Zhang$^2$\\
$^1$The Chinese University of Hong Kong\\
$^2$Tencent AI Lab\\
$^3$Noah's Ark Lab, Huawei Technologies\\
\{yliao, pjli, wlam\}@se.cuhk.edu.hk \\
\{lyndonbing, shumingshi, bradymzhang\}@tencent.com
}

\begin{document}
\maketitle
\begin{abstract}
We propose the task of Quantifiable Sequence Editing (QuaSE): editing an input sequence to generate an output sequence that satisfies a given numerical outcome value measuring a certain property of the sequence, with the requirement of keeping the main content of the input sequence. For example, an input sequence could be a word sequence, such as review sentence and advertisement text. For a review sentence, the outcome could be the review rating; for an advertisement, the outcome could be the click-through rate. The major challenge in performing QuaSE is how to perceive the outcome-related wordings, and only edit them to change the outcome. In this paper, the proposed framework contains two latent factors, namely, outcome factor and content factor, disentangled from the input sentence to allow convenient editing to change the outcome and keep the content. Our framework explores the pseudo-parallel sentences by modeling their content similarity and outcome differences to enable a better disentanglement of the latent factors, which allows generating an output to better satisfy the desired outcome and keep the content. The dual reconstruction structure further enhances the capability of generating expected output by exploiting the couplings of latent factors of pseudo-parallel sentences. For evaluation, we prepared a dataset of Yelp review sentences with the ratings as outcome. Extensive experimental results are reported and discussed to elaborate the peculiarities of our framework. \footnote{\label{fn:code}Our code and data are available at \url{https://bitbucket.org/leoeaton/quase/src/master/}}

\end{abstract}

\section{Introduction}
Typical neural text generation is observed suffering from the problems of repetitions in word n-grams, producing monotonous language, and generating short common sentences \cite{DBLP:conf/emnlp/LiMSJRJ17}. To solve these problems, some researchers branch out into the way of post-editing (could be under some guidance, say sentiment polarity) a given message to generate text of better quality. For example, skeleton-based text generation first outlines a skeleton in the form of phrases/words, and then starts from the skeleton to generate text \cite{DBLP:conf/cvpr/WangLSCC17,Xiao:2016:SST:3016100.3016301}. Another line of works conduct editing on an existing sentence and expect that the output will serve particular purposes better \cite{guu2018edit}. Similarly in conversation, some systems post-edit the retrieval results to generate new sentences as the response \cite{DBLP:journals/corr/SongYLZZ16}. The third type is to perform editing on the input under the guidance of specific style. For example,  \citet{DBLP:conf/nips/ShenLBJ17} take a sentence with negative sentiment as input, and edit it to transfer its sentiment polarity into positive. 

In this paper, we generalize the third type of post-editing into a more general scenario, named \textbf{Qua}ntifiable \textbf{S}equence \textbf{E}diting (\textbf{QuaSE}). 
Specifically, in the training stage, each input sentence is associated with a numeric outcome. For example, the outcome of a review sentence is its rating, ranging from 1 to 5; the outcome of each advertisement is its click-through rate. 
In the test stage, given an input sentence and a specified outcome target, a model needs to edit the input to generate a new sentence that will satisfy the outcome target with high probability. Meanwhile, the output sentence should keep the content described by the input.
For example, given the input sentence ``{The food is terrible}'', a desired output sentence could be ``{The food is OK}'' under the expected outcome ``3.1'' (a neutral sentiment), and  ``{The food is delicious}'' under the expected outcome ``4.0''.
If no outcome target is given, the model could generate ``{The food is extremely delicious}'', by defaulting the best outcome, or ``{The food is extremely terrible}'', by defaulting the worst outcome. 

Our problem setting is more general than previous works in two major aspects: (1) The outcome here is numerical, and it can be regarded as a generalization of the categorical outcome in  \cite{DBLP:conf/nips/ShenLBJ17,DBLP:conf/icml/HuYLSX17,DBLP:journals/corr/abs-1807-03586}. With such numerical outcome, it is impossible to construct two corpora as counterpart of each other as done in \cite{DBLP:conf/nips/ShenLBJ17,DBLP:journals/corr/abs-1807-03586}. 
(2) The editing operation is under a quantifiable guidance, i.e. the specified outcome or the defaulted extrema. For example,
we can specify a particular target rating, such as 3.1 or 4.0, as the expected outcome. Although \citet{pmlrv70mueller17a} also take outcome-associated sentences for training, their model does not perform such outcome-guided editing for sentence generation.

Considering that the goal of the task is to generate an output that satisfies a specified outcome and keeps the content unchanged, QuaSE is challenging in a few aspects. Firstly, a model should be able to perceive the association between an outcome and its relevant wordings. For the previous example ``{The food is terrible}'', the model needs to figure out that the low rating is indicated by the word ``{terrible}'', instead of ``{food}''.
Secondly, when performing editing, the model should keep the content, and only edits the outcome-related wordings. Moreover, the model needs to take a specified outcome into account and generate an output that satisfies the specified outcome value with high probability. Continuing the running example, given the expected outcome 3.1, ``{The food is OK}'' is an appropriate output, but ``The food is extremely delicious'' and ``{The service is OK}'' are not. Thirdly, we do not have readily available data, such as data points like [input sentence: ``The food is terrible'', expected outcome: 4.0, output sentence: ``The food is delicious''] to show the model what the revised output should look like, that meet our need to train models. 

We propose a framework to address this task. The fundamental module of our framework is a Variational Autoencoder (VAE) \cite{DBLP:journals/corr/KingmaW13} to encode each input sentence into a latent content factor and a latent outcome factor, capturing the content and the outcome related wordings respectively. 
We propose to leverage pseudo-parallel sentence pairs (i.e, the two sentences in a pair have the same or very similar content, but different outcome values) to enhance our model's capability of disentangling the two factors, which allows attributing the wording difference of the sentences in a pair to the outcome factor, and the wording similarity to the content factor. For sentence reconstruction, we employ a Recurrent Neural Network (RNN) based decoder \cite{DBLP:conf/nips/SutskeverVL14} that takes as input the combination of a content factor and an outcome factor. To further enhance the capability of generating expected output, we introduce a dual reconstruction structure which exploits the couplings of latent factors of pseudo-parallel sentences. Specifically, it attempts to reconstruct one sentence in a pair from the combination of its outcome factor and the other sentence's content factor, based on the intuition that the wording difference in a pair is outcome-related. 
In the test stage, taking a sentence and a specified outcome target as input, our model generates a revised sentence which likely satisfies the specified target, and meanwhile the content is preserved as much as possible. 

To evaluate the efficacy of our framework, we prepared a dataset of Yelp review sentences with the ratings as outcome. Compared with state-of-the-art methods handling similar tasks, experimental results show that our framework can generate more accurate revisions to satisfy the target outcome and transfer the sentiment polarity, meanwhile it keeps the original content better. Ablation studies illustrate the effectiveness of the designed components for enhancing the performance. We have released the prepared dataset and the code of our model to facilitate other researchers to do further research along this line, refer to Footnote $\ref{fn:code}$.


\section{Model Description}
\subsection{Problem Setting and Model Overview}
In the task of Quantifiable Sequence Editing (QuaSE), the aim is to edit an input sentence $X_0$ under the guidance of an expected outcome value $R^*$ to generate a new sentence $X^*$ that will satisfy $R^*$ with high probability. For training a model, we are given a set of sentence-outcome tuples ($X$, $R$).

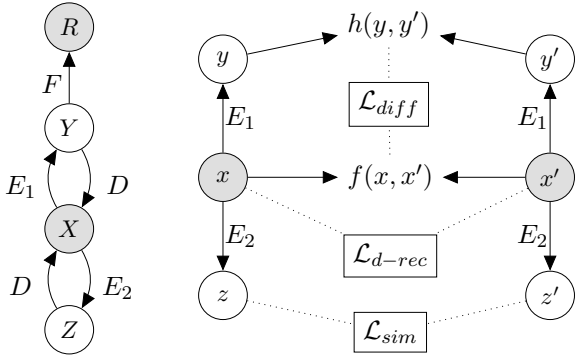
\begin{figure}[t]
    \centering
    \resizebox{1\linewidth}{!}{\input{model_large.tex}}
    \vspace{-0.4cm} 
    \caption{Model Overview. }
    \label{model}
    \vspace{-0.54cm}
\end{figure}

Our proposed model for training is depicted in Figure \ref{model}. 
The left hand side models individual sentences. Specifically, it employs two encoders, i.e. $E_1$ and $E_2$, to encode a single sentence $X$ into two latent factors $Y$ and $Z$ which capture the outcome and content properties respectively. In contrast, \citet{pmlrv70mueller17a} employ a single factor for capturing these two properties, which limits the capability of  distinguishing one property from the other. As a consequence, when editing a sentence towards a given outcome, the sentence content is likely to be changed, which should be suppressed as much as possible.
An RNN-based decoder $D$ takes the concatenation of $Y$ and $Z$ to reconstruct the input $X$. Moreover, a transformation function $F$ predicts $R$ with $Y$.

The right hand side models pseudo-parallel sentence pairs (automatically generated from the above tuples), so we first introduce the concept of pseudo-parallel sentences as follows. Let ($x$, $x'$) denote a pair of pseudo-parallel sentences, $x$ and $x'$ should describe the same or similar content, but their outcomes are different. Note that we use lowercase letters to denote variables related to sentence pairs for better clarity. For two sentences in a pair, the difference of their outcome factors  $h(y,y')$ is attributed to their wording difference $f(x,x')$, resulting in the loss $\mathcal{L}_{diff}$; the similar contents of two sentences should result in similar content factors, i.e. minimizing the loss $\mathcal{L}_{sim}$; moreover, a dual reconstruction loss $\mathcal{L}_{d-rec}$ is minimized to enhance the capability of generating expected output. 

Overall, the model minimizes the losses from modeling single sentences and sentence pairs.
After the model is trained, a separated component is applied for editing an input sentence to output a revision that satisfies a specified outcome target.

\subsection{Modeling Single Sentences}\label{modelsentences}
In probabilistic theory, we need to maximize the log-likelihood of observing the training sentence-outcome tuples ($X$,$R$), denoted as follows:
\begin{equation} \label{full obs}
\begin{split}
    \log \int p(X,R) = &\log \int p(X|Y,Z)p(Y,Z)dYdZ \\
    &+ \log \int p(R|Y)p(Y)dY
\end{split}
\end{equation}
However, the integration in the first term on the right hand side is intractable. Inspired by the idea of VAE \cite{DBLP:journals/corr/KingmaW13}, we alternatively maximize the Evidence Lower Bound (ELBO) \cite{DBLP:journals/corr/BleiKM16} incorporating variational distributions, i.e. $q(Y|X)$ and $q(Z|X)$. Thus, this term is approximated as follows:
\begin{equation} \label{1EBLO}
\begin{split}
    &\log \int p(X|Y,Z)p(Y,Z)dYdZ \geq -[\mathcal L_{rec}+\mathcal L_{kl}]\\
        &\mathcal L_{rec} = -\mathbb{E}_{Y,Z\sim q(Y|X),q(Z|X)}[\log p(X|Y,Z)]\\
      &\mathcal L_{kl} = KL[q(Y|X)|p(Y)] +KL[q(Z|X)|p(Z)]
\end{split}
\end{equation}
where, the term $\mathcal L_{rec}$ denotes the error of reconstructing $X$. As advocated by \cite{DBLP:journals/corr/KingmaW13} and \cite{DBLP:conf/conll/BowmanVVDJB16}, the variational distributions $q(Y|X)$ and $q(Z|X)$ are modelled as Gaussian distributions, i.e. $q(Y|X) = \mathcal G(\mu_{Y|X}, \sigma_{Y|X})$, and $q(Z|X) = \mathcal G(\mu_{Z|X},\sigma_{Z|X})$. The expectation $\mathbb E(\cdot)$ can be efficiently approximated using one Monte-Carlo sample, for example, $Y \sim q(Y|X)$ and $Z \sim q(Z|X)$. In practise, we can alternatively employ $Y=\mu_{Y|X} $ and $Z=\mu_{Z|X}$ instead of sampling since they are the means of the Gaussian distributions. We employ two encoder networks $E_1$ and $E_2$ to generate $\mu_{Y|X}$ and $\mu_{Z|X}$ respectively from the sentence $X$, i.e. $\mu_{Y|X}=E_1(X)$ and $\mu_{Z|X}=E_2(X)$.  $p(X|Y,Z)$ is the probability of observing the sentence $X$ given $Y$ and $Z$, which is modelled by a decoder network $D$. Thus, the reconstruction loss can be rewritten as:
\begin{equation}
    \mathcal L_{rec} = H(X,D(E_1(X),E_2(X)))
\end{equation}
\noindent where $H$ is the cross entropy loss for the decoder.

The term $\mathcal L_{kl}$ in Equation \ref{1EBLO} denotes the KL-divergence between the variational posterior distribution and the prior distribution. Following previous works \cite{pmlrv70mueller17a}, the priors $p(Y)$ and $p(Z)$ are defined as a zero-mean Gaussian distribution, i.e. $p(Y) = p(Z)=\mathcal G(\textbf{0},\textbf{I})$. The loss $\mathcal L_{kl}$ serves as a regularization term enforcing that the variational posterior distribution resembles the prior distribution, which also avoids overfitting.


The second term in Equation \ref{full obs} models the log-likelihood of the outcomes. We adopt the usually used Taylor approximation for the calculation, where this term is approximated by an affine transformation from the outcome factor $Y$ to the outcome $R$, denoted as $F(Y)$. Then, we define the loss as the square error between $R$ and $F(Y)$:
\begin{equation}
\mathcal L_{mse} = (R-F(Y))^2
\end{equation}

Although \citet{pmlrv70mueller17a} also model individual sentences and their outcomes, in their model, each sentence is only encoded into one latent factor to capture both outcome and content properties. In contrast, we disentangle two latent factors from a single sentence to model the outcome and the content separately to provide more flexibility. Moreover, such design allows the incorporation of the pseudo-parallel sentences, which will be described in the next subsection.

\subsection{Exploiting Pseudo-Parallel Sentences}

As mentioned above, pseudo-parallel sentences are similar in terms of the content but different in terms of the outcome. E.g., Table \ref{tab:parallel-sentences} 
shows a pair of pseudo-parallel sentences, where both talk about ``the restaurant'', but with different sentiments (i.e. ratings). For the pair ($x$, $x'$), let $y$ and $y'$ denote their outcome factors, $z$ and $z'$ denote their content factors.
We design three components to leverage pseudo-parallel sentences to enhance our model's capabilities of disentangling the two types of factors and generating the desired output sentences.

\begin{table}[h]
\small
    \centering
    \begin{tabular}{@{}c@{~}|@{~}l@{}}
    \hline
         $x$& I will never come back to the restaurant. \\ \hline
         $x'$& I will definitely come back to the restaurant, recommend! \\\hline
    \end{tabular}
    \caption{A pair of pseudo-parallel sentences.}
    \label{tab:parallel-sentences}
\end{table}

\subsubsection{Modeling Outcome Difference}
We exploit the wording difference $f(x,x')$ between $x$ and $x'$. Note that the preparation (discussed in Section \ref{sec:data_pre}) determines that a pair of pseudo-parallel sentences are very likely to differ in the outcome factors, denoted as $h(y,y')$. 
Thus, by aligning the surface wording difference of two sentences in a pair and the difference in their outcome factors, we intend to improve the performance of the encoder $E_1$ for generating the outcome factor.
$f(x, x')$ and $h(y,y')$ are defined as follows:
\begin{equation} \label{diff}
\begin{split}
    f(x,x') = & inc(x,x')\oplus dec(x,x') \\
    h(y,y') = & y-y' = E_1(x) - E_1(x')
\end{split}
\end{equation}
\noindent where $inc(x,x')$ and $dec(x,x')$ are embeddings capturing the wording difference between $x$ and $x'$. $inc(x,x')$ denotes the ``increment" from $x$ to $x'$, i.e. the terms that appear in $x'$ but not in $x$. $dec(x,x')$ denotes the ``decrement". If there are multiple terms in the difference, we sample one term for $inc$ or $dec$. For the example in Table \ref{tab:parallel-sentences}, $dec(x,x')$ is the embedding of ``never", and $inc(x,x')$ could be the embedding of ``definitely" or ``recommend". The effect of outliers during sampling anneals since the training data contain sufficient pairs of sentences. The symbol $\oplus$ denotes concatenation. $h(y,y')$ is defined as the subtraction between the outcome factors.

We employ a regression network $U$ to align $f(x,x')$ and $h(y,y')$, and the loss $\mathcal L_{diff}$ is: 
\begin{equation}
\mathcal L_{diff} = ||h(y,y')-U[f(x,x')]||^2
\end{equation}

\subsubsection{Modeling Content Similarity}
Another property of two pseudo-parallel sentences is that they share similar content. To capture it, we design a loss function minimizing the square error between the content factors. 
\begin{equation}
    \mathcal L_{sim} =||z-z'||^2  = ||E_2(x)-E_2(x')||^2
\end{equation}
Minimizing $\mathcal L_{sim}$ helps the encoder $E_2$ generate the content factor more accurately.

\subsubsection{Dual Reconstruction}
The decoder $D$ is not only used in Section \ref{modelsentences} to reconstruct a single training sentence, but also employed for generating output sentences in the test stage (Section \ref{sec:revise}). To improve the robustness of $D$, we propose a dual reconstruction component based on the pseudo-parallel sentences. Different from reconstructing an original sentence in Section \ref{modelsentences}, in the dual reconstruction, given a sentence $x$, we reconstruct its dual sentence $x'$. 

Specifically, we first encode $x$ and $x'$ into their 
outcome factors $y/y'$ and content factors $z/z'$. Since $x$ shares similar content with $x'$, its content factor $z$, when combined with the outcome factor $y'$ of $x'$, should nearly reconstruct $x'$. For such dual reconstruction, the loss is written as:
\begin{equation}
\begin{split}
    \mathcal L^{d-rec}_{x'; x} =&  H(x',D(E_1(x'),E_2(x)))\\
    = & H(x',D(y',z))
\end{split}
\end{equation}
The same dual reconstruction process applies to the counterpart of $x'$, i.e. $x$. 
Thus, the whole dual reconstruction loss is as follows:
\begin{equation}\label{dual-rec}
    \mathcal L_{d-rec} = \mathcal L^{d-rec}_{x'; x} + \mathcal L^{d-rec}_{x; x'}
\end{equation}
\noindent Note that the encoders $E_1/E_2$ and the decoder $D$ here refer to exactly the same networks (i.e., the parameters are shared) as used in Section \ref{modelsentences}.

The specific design of the networks are as follows. $E_1/E_2$: RNNs of GRUs with a fully connected neural network appended to the last state to add some noise, which is a reparameterization alternative for sampling. Their outputs are the outcome and content factors, respectively. $D$: An RNN of GRU cells. The RNN takes the concatenation of an outcome factor and a content factor as the initial state for decoding. $F$: A fully connected network. It takes an outcome factor as input and outputs an outcome value. $U$: A fully connected network. It takes $f(x,x')$ as input to predict $h(y,y')$.




\subsection{Joint Training}
Considering all the aforementioned components, we define a joint loss function as:
\begin{equation} \label{jointloss}
\begin{split}
    \mathcal L_{joint} = &\lambda_{rec}\mathcal L_{rec} + \lambda_{kl} \mathcal L_{kl} + \lambda_{mse}\mathcal L_{mse} + \\ &\lambda_{diff}\mathcal L_{diff} +
    \lambda_{sim}\mathcal L_{sim} + \lambda_{d-rec}\mathcal L_{d-rec}
\end{split}
\end{equation}
in which each component is associated with a weight. Following the sigmoid annealing schedule \cite{DBLP:conf/conll/BowmanVVDJB16}, we design the following strategy to tune the weights: (1) Tune the weights $\lambda_{rec}$ and $\lambda_{mse}$ on the validation dataset under the metric MAE (refer to Section \ref{sec:metric}), while fixing the other weights to zeros. We set $\lambda_{rec} + \lambda_{mse}=1$; (2) Fixed the weights tuned in the first step. For each remaining loss, gradually increase the weight from 0 to 1 during the training, until the reconstruction loss $\mathcal L_{rec}$ or the outcome prediction loss $\mathcal L_{mse}$ becomes worse. 
The strategy prioritizes $\mathcal L_{rec}$ and  $\mathcal L_{mse}$, since they are the core components for generating the revised sentences.


\section{Editing under Quantifiable Guidance}
\label{sec:revise}
In the test, the trained model edits an input sentence $X_0$ and outputs a revision $X^*$ that is likely to satisfy the specified outcome target $R^*$, and meanwhile preserves the content as much as possible. 

We first encode $X_0$ with $E_1$ and $E_2$ to get $Y_0$ and $Z_0$ respectively. The next step is to modify $Y_0$ to get a new outcome factor $Y^*$ that is likely to generate the target outcome $R^*$. 
The process to determine a suitable $Y^*$ is as follows. We first assume $Y$ follows the Gaussian distribution $Y \sim \mathcal G(Y_0 = E_1(X_0),\sigma)$, the mean of which is $Y_0$. Then we choose  $\mathcal C = \{Y: \mathcal G(Y|E_1(X_0),\sigma) > \tau \}$ as the feasible range for $Y^*$, where $\tau$ is a threshold. $\mathcal C$ will expand if $\tau$ is set smaller, and thus allowing more revisions. Finally, $Y^*$ is determined as follows:
\begin{equation}
    Y^* = \argmin_{Y \in \mathcal C} (F(Y) - R^*)^2
\end{equation}
Note that in \cite{pmlrv70mueller17a}, $Y^*$ is determined as $\argmax_{Y \in \mathcal C} F(Y)$, which does not consider an outcome target.
The revised sentence $X^*$ is generated from $X_0$ and $Y^*$ via the decoder $D$:
\begin{equation}
    X^* = D(Y^*,Z_0)
\end{equation}
Thus, the content of $X_0$ is preserved with $Z_0$, and the expected outcome is achieved with $Y^*$.

\section{Experiments} \label{sec:experiment}
\subsection{Dataset Preparation}
\label{sec:data_pre}
Our dataset contains sentences extracted from Yelp reviews \footnote{\mbox{\small{\url{https://www.yelp.com/dataset/challenge}}}}, where each review is associated with a rating in \{1, 2, 3, 4, 5\}. Specifically, we employ the sentences with sentiment polarity (i.e. positive or negative) used in \cite{DBLP:conf/nips/ShenLBJ17} as the primary portion of our data. After some cleaning, we obtain about 520K sentences. To add neutral sentences, we randomly select 80K sentences from the original reviews with neutral sentiment (i.e. rating 3). To make sure that the neural sentences added by us are describing the same domain, we only pick neural sentences whose tokens are all in the vocabulary of the primary data. The vocabulary size of the dataset is 9,625. In total, our dataset contains 599K sentences, and we randomly hold 50K for test, 10K for validation, and the remaining for training. 

For training, we need each input sentence being associated with a rating value, and for test, we need to measure the rating of a generated sentence to check if the generated sentence satisfies the specified outcome target. Therefore, an automatic method is needed for measuring the rating values of training sentences and generated sentences. We employ the sentiment analyzer in Stanford CoreNLP \cite{manning-EtAl:2014:P14-5} to do so. Specifically, we first invoke CoreNLP to output the probability of each rating in \{1, 2, 3, 4, 5\} for a sentence, then we take the sum of the probability-multiplied ratings as the sentence rating. 
Some statistics of the data is given in Table \ref{tab:statistics}. Hereafter, we use ``rating'' and ``outcome'' interchangeably.

\begin{table}[t]
\small
    \centering
    \begin{tabular}{c|cccc}
    \hline
         Rating interval &$[1,2)$&$[2,3)$&$[3,4)$&$[4,5]$  \\ \hline
         Sentence\# &$34273$ &$231740$&$165159$&$167803$ \\ \hline
    \end{tabular}
    \caption{Numbers of sentences in each rating interval.}
    \label{tab:statistics}
\end{table}

One may think that would it be possible to use the original rating given by Yelp users as outcome for training? We did not use it for two reasons: (1) We want the ratings of training sentences and generated sentences are measured with a consistent method; (2) In fact, we find that the predicted rating with CoreNLP has a Pearson correlation of 0.85 with the rating given by users. Note that the original Yelp data only has ratings for entire reviews. We derived the sentence ratings by users like this: a sentence takes as its rating the average of the ratings of those reviews where it appears in. Human evaluation in \cite{DBLP:conf/nips/ShenLBJ17} shows that a similar method for deriving polarity is basically reasonable as well. 

\begin{table*}[t]
    \centering
    \begin{tabular}{@{}c||c|c|c|c|c||c|c|c|c|c@{}}
    \hline
         \multirow{2}{*}{~}&\multicolumn{5}{|c||}{MAE}&\multicolumn{5}{|c}{Edit Distance}\\ \cline{2-11}
         &T=1&T=2&T=3&T=4&T=5&T=1&T=2&T=3&T=4&T=5 \\ \hline
         Original &2.2182&1.2379&0.8259&0.9279&1.7818& N/A&N/A&N/A&N/A&N/A\\ \hline
         S2BS&1.6839&0.9444&0.7567&0.7572&1.3024&6.6439&5.342&4.9390&5.005&6.2290 \\ \hline
         Our Model&1.4162&0.6298&0.7408&0.5377&0.9408&7.9191&4.7&3.4505&4.13&8.0094\\ \hline
    \end{tabular}
    \caption{MAE and Edit Distance for our proposed model and S2BS. T refers to the target outcome.}
    \label{basic_comp}
\end{table*}

For preparing the pseudo-parallel sentences, we first follow the ideas in \cite{guu2018edit} to generate some initial pairs. Specifically, we first calculate the Jaccard Index (JI) for each pair of sentences, and keep those with JI values no less then 0.5 as the initial pairs. Note that such initial pairs could contain many false positives (roughly 10\% as manually evaluated on the Yelp corpus in \cite{guu2018edit}), because the JI calculation does not distinguish content words and outcome words. To solve this problem, we add another constraint: for an initial pair to be regarded as a pseudo-parallel sentence pair, the difference of the two sentences' ratings should be no less than 2. Here, the idea is that given the two sentences are similar enough in wordings (JI $\geq$ 0.5), if their rating scores are dissimilar enough, it looks plausible to conjecture that their wording difference is more likely outcome-related and causes the rating difference. In fact, such wording difference is exactly what we want to capture with pseudo-parallel sentence pairs. In total, we obtain about 604K sentence pairs from the single training sentences. For conducting the joint training with both single sentences and pseudo-parallel pairs, we make each data point composed of a single sentence and a sentence pair. To do so, we couple each sentence pair with a single sentence, thus we can use all pairs for training. Note that because we have more sentence pairs, some single sentences are used twice randomly in composing the data points.



\subsection{Comparative Methods}
Our model is compared with two state-of-the-art models handling similar tasks. 

\textbf{Sequence to Better Sequence (S2BS)} \cite{pmlrv70mueller17a}:
For training, S2BS also requires each sentence is associated with an outcome. For test, S2BS only revises a sentence such that the output is associated with a higher outcome, which is not a specified value. For comparison, we adapt our revision method for S2BS, by which their trained model is able to conduct quantifiable sentence revision. We tune the parameters for S2BS by following the suggestions in their source code. 


\textbf{Text Style Transfer (TST)} \cite{DBLP:conf/nips/ShenLBJ17}: 
In TST, the sentiment of each sentence is labelled as negative or positive. The model is able to revise a negative sentence into positive, or vice versa. Their task can be treated as a special case of our QuaSE task: we set the outcome target to 1 for the input sentences that are associated with outcomes larger than the neutral rating 3, thus, our task is equal to revising a positive sentence into negative. We follow the suggested parameters reported in \cite{DBLP:conf/nips/ShenLBJ17}. 

\subsection{Evaluation Metric and Parameter~Setting}
\label{sec:metric}
Considering that our model's task is to revise a sentence such that its outcome (predicted by Stanford CoreNLP) satisfies a specified target, we define the metric as the mean absolute error (MAE) between the specified target outcome and the outcomes of revised sentences.
\begin{equation}
    MAE = \frac{1}{|S|}\sum_{X_i\in S} |R_i-R^*|
\end{equation}
\noindent where $S$ is the set of revised sentences $X_i$, $R^*$ is the target outcome, and $R_i$ is the outcome of $X_i$.

After tuning on the validation set, the determined parameters are: $\lambda_{rec} = 0.75$, $\lambda_{kl} = 0.6$, $\lambda_{mse} = 0.25$, $\lambda_{diff} = 0.2$, $\lambda_{sim} = 0.2$, $\lambda_{d-rec} = 0.1$, and the dimensions of the two factors are both 50.
The parameter $\tau$ for revision takes $\exp(-100000)$ for both our model and S2BS.

\subsection{Automatic Evaluation}
We compare our model with S2BS by specifying five target ratings, namely 1, 2, 3, 4, and 5. Both our model and S2BS are fed the sentences in the testing dataset. For each sentence, both models are required to generate five revised sentences, each satisfying one of the target ratings. We evaluate the MAE between the target outcome and the outcome of the revised sentences. Each model is trained for three times and the average results are reported in Table \ref{basic_comp}. ``Original" refers to the MAE between the targets and the ratings of input sentences. We can observe that the MAE values of both our model and S2BS are smaller than Original. It demonstrates that both models are able to revise the sentences towards the outcome targets. Furthermore, compared with S2BS, our model achieves smaller MAE values. One major reason is that we disentangle a content factor and an outcome factor, and design three components to leverage pseudo-parallel sentences. By modeling the wording difference, our model captures the keywords that cause the difference in the outcome. By enforcing the content factors of pseudo-parallel sentences to be similar, the model is capable to generate the content factor more precisely. Moreover, the dual reconstruction can guide the editing procedure with the hints from the parallel sentences. In contrast, S2BS only disentangles one factor for capturing both content and outcome properties, and thus it cannot perform the same operations on sentence pairs.
The MAE for T=5 is smaller than that for T=1. This is partially due to the fact that the outcomes of the test sentences are closer to 5, refer to Table \ref{tab:statistics}. We also report the average Edit Distance between the input sentences and the generated sentences to measure the degree of revisions. For T=1 and T=5, our model conducts more editing than S2BS, which brings in better MAE, while for T=\{2, 3, 4\}, our model generates more accurate sentences (i.e. better MAE) with less editing. This observation coincides with the fact that we need more editing to revise a sentence towards an extreme target (i.e., 1 and 5), such as including degree adverbs ``very'' and ``extremely''.

\begin{table}[t]
    \centering
    \begin{tabular}{c||c|c}
    \hline
         &Neg. to Pos.&Pos. to Neg.\\ \hline
         TST&0.7280&0.7097\\ \hline
         Our Model&0.8836&0.7862 \\ \hline
    \end{tabular}
    \caption{Accuracy comparison with TST.}
    \label{accuracy}
\end{table}

We also compare our model with TST for sentiment polarity transfer. We employ the same evaluation metric as used in \cite{DBLP:conf/nips/ShenLBJ17}: the sentiment accuracy of the transferred sentences. We define the revised sentences with ratings larger than 3 as positive, smaller than 3 as negative. The results are given in Table \ref{accuracy}, where two accuracy values are reported: negative to positive, and the reverse. The results show that our model achieves better accuracy than TST in both transfer directions. One reason is that our method models the associations between each sentence and its outcome, and thus captures the sentiment wordings better. Our model is far better for transferring negative sentences into positive, moreover, both models achieve better performance for this transfer direction. We can probably attribute the reason to the imbalanced training data: 55\% positive sentences v.s. 45\% negative sentences.

\begin{table}[t]
    \centering
    \begin{tabular}{@{}c@{~}||@{~}c@{~}|@{~}c@{}}
    \hline
         & Content Preservation &  Fluency \\ 
         & (Range: [0, 2]) &   (Range: [1, 4])\\ \hline
         TST & 1.02 & 2.56 \\ \hline
         S2BS & 0.70 & 2.53 \\ \hline
         Our Model & 1.38 & 2.48 \\ \hline
    \end{tabular}
    \caption{Manual evaluation.}
    \label{tab:manual}
\end{table}

\begin{table*}[t]
    \centering
    \begin{tabular}{c|p{12cm}}
    \hline
          & Generated sentence  \\ \hline
         E.g. 1 &this tire center is amazing .\\
         T=1&this tire center is horrible .\\
         T=3&this tire center is really good .\\
         T=5&this tire center is amazing .\\ \hline
         
        E.g. 2 &horrible food !\\
         T=1&horrendous \\
         T=3&their food amazing !\\
         T=5&amazing delicious food ! recommend ! \\ \hline
         
         E.g. 3 &decent food and wine selection , but nothing i will rush back for .\\
         T=1&decent food and wine selection , but nothing i will rush for no .\\
         T=3&decent food and wine selection , but i will never look back for .\\
         T=5&decent food and wine selection , but excellent service, will return !\\ \hline

         E.g. 4 &our first time and we had a great meal , wonderful service .\\
         T=1&our first time and we had a terrible meal , stale service .\\
         T=3&our first time and we had a great meal , we have service .\\
         T=5&our first time and we had a great meal , wonderful service .\\\hline
         
         E.g. 5 &food is very addiction tasty !\\
         T=1&food is just horrible here ?\\
         T=3&food is just addiction here !\\
         T=5&food is very yummy addiction !\\ \hline
    \end{tabular}
    \caption{Case study.}
    \label{tab:cases}
\end{table*}

\subsection{Manual Evaluation}
We hire five workers to manually evaluate the quality of 500 sentences generated by each of our model and the compared models. The result is shown in Table \ref{tab:manual}. ``Content Preservation" measures whether the generated sentence preserves the content of the input sentence. The score range is \{2: fully preserved, 1: partially preserved, 0: not preserved\}. 
``Content Preservation" is an important metric in this task since it is required that the output sentence and the original sentence should describe the same content subject. ``Fluency" measures the grammatical quality of a sentence, which ranges from 1 (bad) to 4 (good), by following the definition in TST~\cite{DBLP:conf/nips/ShenLBJ17}.

The result shows that our model achieves the best content preservation score. Our editing procedure explicitly fixes the content factor and only modifies the outcome factor, which helps better preserve the content. In contrast, S2BS and TST include only one shared factor for both the content and the outcome, thus fail to distinguish one from the other.
For the ``Fluency" metric, S2BS and TST are slightly better than our model. Generally speaking, it is because our model introduces more powerful components for modeling the outcome differences between pseudo-parallel sentences, so as to achieve our goal of editing an input sentence to satisfy the expected outcome. However, these components do not contribute to the language quality of generated sentences.

\subsection{Case Study}
We show some examples produced by our model in Table \ref{tab:cases}. For each input, we specify three targets: 1, 3, and 5. 
For the first and the forth examples, the original sentences are not revised when the target rating is set to 5 (i.e., T=5) since the original sentences are already quite positive. For the first example, when T=3, ``amazing" is revised to a relatively less positive phrase ``really good". This case demonstrates that our model is able to capture the subtle difference in word sentiments, so that it can revise sentences reasonably according to the quantifiable rating guidance.
Moreover, for the second example, we notice that our model revises the original sentence ``horrible food !" to ``amazing delicious food ! recommend !" for T=5. This case shows that our model not only changes one word with another having different sentiment, e.g. ``horrible" to ``amazing delicious", but also creatively introduces words from a new perspective, e.g. ``recommend".

\subsection{\mbox{Ablation and Tuning Behavior Discussions}}\label{sec:ablation}
Recall that our model is a combination of a revised VAE, which disentangles two factors from a sentence to enable the subsequent design, and a coupling component modeling pseudo-parallel sentence pairs. For the three losses of the coupling component, we show their effects under the MAE metric in Table \ref{tab:ablation}. 
``None" refers to all three losses are removed, and it is basically worse than S2BS, which implies only using the revised VAE does not work well. As more losses added, the performance is gradually improved. Moreover, the dual reconstruction is more effective than the others. 

In the weight tuning, the first step only tunes the weights of $\mathcal L_{rec}$ and $\mathcal L_{mse}$. 
We observe that solely minimizing $\mathcal L_{rec}$ and $\mathcal L_{mse}$ also decreases $\mathcal L_{sim}$, because in this process, the encoder $E_2$ becomes more capable of disentangling the content factor, and thus $z$ and $z'$ become similar as they come from two similar input sentences, i.e. pseudo-parallel sentences. 
Another observation is that solely minimizing $\mathcal L_{rec}$ and $\mathcal L_{mse}$ increases $\mathcal L_{d-rec}$ after some training epochs. To analyze the reason, let us assume there is a sentence $x$ in the training set. Thus, the loss of reconstructing $x$ from $y$ and $z$ is included in $\mathcal L_{rec}$. Assume that $x$ is also included in a pseudo-parallel pair, and thus the loss of reconstructing $x$ from $y$ and $z'$ is included in $\mathcal L_{d-rec}$. The only difference between the two losses lies in the content factors $z$ and $z'$. Given that $z$ and $z'$ are not enforced to resemble each other when $\mathcal L_{sim}$ is excluded from this tuning step, $\mathcal L_{rec}$ and $\mathcal L_{d-rec}$ cannot be minimized simultaneously. Moreover, when we minimize $\mathcal L_{sim}$ in the second step with the weights of $\mathcal L_{rec}$ and $\mathcal L_{mse}$ fixed, we observe that $\mathcal L_{d-rec}$ also decreases, which complies with the above analysis.

\begin{table}[t]
    \centering
    \begin{tabular}{c|c|c|c}
    \hline
         &T=1&T=3&T=5 \\ \hline \hline
         S2BS& 1.6839 & 0.7567 & 1.3024\\ \hline
         
         None& 1.6639& 0.7684&1.5434\\ \hline
         
         $\mathcal L_{sim}$ &1.6090& 0.8258&1.5233 \\ \hline
         $\mathcal L_{diff}$ &1.6793 &0.8017 &1.3140\\ \hline
         $\mathcal L_{d-rec}$&1.5191 & 0.7784 &1.1218\\ \hline
         $\mathcal L_{sim}, \mathcal L_{diff}$&1.4991 & 0.8218&1.3705\\ \hline
         $\mathcal L_{sim}, \mathcal L_{d-rec}$& 1.4101& 0.8027&1.1246\\ \hline
         $\mathcal L_{diff}, \mathcal L_{d-rec}$& 1.3879& 0.7786&1.1413\\  \hline
         ALL & 1.4162 & 0.7408 & 0.9408\\  \hline
         
    \end{tabular}
    \caption{Ablation study.}
    \label{tab:ablation}
\end{table}


  
  
  
  

\section{Related Works}



Inspired by the task of image style transfer \cite{gatys2016image,DBLP:conf/nips/LiuT16}, researchers proposed the task of text style transfer and obtained some encouraging results \cite{aaai2018fustyle,DBLP:conf/icml/HuYLSX17,DBLP:journals/corr/JhamtaniGHN17,DBLP:journals/corr/abs-1711-09395,yezhang,Juncen_Li_naacl2018,Prabhumoye_acl2018,Niu2018tacl}.
Existing studies on text style transfer mainly aim at transferring text from an original style into a target style,  e.g., from negative to positive, from male to female, from rude/normal to polite; from modern text to Shakespeare style, etc. In contrast, our proposed task QuaSE assumes each sentence is associated with an outcome pertaining to continues values, and the editing is under the guidance of a specific target. 

To transfer the style of a sentence, the paradigm of most works \cite{DBLP:conf/nips/ShenLBJ17,pmlrv70mueller17a,Prabhumoye_acl2018} first learns the latent representation of the original sentence and then applies a decoder to generate the transferred sentence. A line of works \cite{DBLP:conf/nips/ShenLBJ17,pmlrv70mueller17a}, including the studied task in this paper, assume that only non-parallel data is available for training. In such settings, VAEs \cite{DBLP:journals/corr/KingmaW13} are employed to learn the latent representations of sentences. \citet{DBLP:conf/nips/ShenLBJ17} assume a shared latent content distribution across text corpora belonging to different styles, and leverages the alignment of latent representations from different styles to perform style transfer. \citet{pmlrv70mueller17a} associate the latent representations with a numerical outcome, which is a measurement of the style. A transferred sentence is generated from a modified latent representation. Different from the aforementioned works based on latent representations, \citet{Juncen_Li_naacl2018} propose a simpler method that achieves attribute transfer by changing a few attribute marker words or phrases in the sentence that are indicative of a particular attribute, while leaving the rest of the sentence largely unchanged. The simple method is able to generate better-quality sentences than the aforementioned works. Besides style transfer, sentence editing models can be developed for other tasks. For example, \citet{Schmaltz2017Adapting} propose neural sequence-labelling models for correcting the grammatical errors of sentences.

\section{Conclusions}
We proposed a new task namely Quantifiable Sequence Editing (QuaSE), where a model needs to edit an input sentences towards the direction of a numerical outcome target. To tackle this task, we proposed a novel framework that simultaneously exploits the single sentences and pseudo-parallel sentence pairs. For evaluation, we prepared a dataset with Yelp sentences and their ratings. Experimental results show that our framework outperforms the compared methods under the measures of sentiment polarity accuracy and target value errors. Case studies show that our framework can generate some interesting sentences.

\bibliographystyle{acl_natbib}
\bibliography{emnlp2018}
\end{document}

%% file: model_large.tex
\begin{tikzpicture}

    \node[obs]  (r) {$R$};
    \node[latent, below = 0.75of r] (y){$Y$};
    \node[obs, below = 0.75of y] (x) {$X$};
    \node[latent, below = 0.75of x] (z) {$Z$};
	
	\edge[left = 30]{y}{r};
	\edge[bend left = 30]{y}{x};
	\edge[bend left = 30]{x}{y};
	\edge[bend left = 30]{x}{z};
	\edge[bend left = 30]{z}{x};

    \node[latent, right = 1.5of y, yshift = 1 cm] (Y){$y$};
    \node[obs, below = of Y] (X) {$x$};
    \node[latent, below = of X] (Z) {$z$};
	
	\edge[left = 30]{X}{Y};
	\edge[left = 30]{X}{Z};
	
    \node[latent, right = 4 of Y] (Y2){$y'$};
    \node[obs, below = of Y2] (X2) {$x'$};
    \node[latent, below = of X2] (Z2) {$z'$};
	
	\edge[left = 30]{X2}{Y2};
	\edge[left = 30]{X2}{Z2};
	
	\node[draw = none, fill = none,  right = 1.3of X] (Dxx){$f(x,x')$};
	\node[draw = none, fill= none, right = 1.3of Y,yshift = 0.5 cm] (Dyy) {$h(y,y')$};
	\node[draw = none, fill = none, below = 0.2of y, xshift = -0.7 cm]{$E_1$};
	\node[draw = none, fill = none, below = 0.2of y, xshift = 0.7 cm]{$D$};
	\node[draw = none, fill = none, below = 0.2of r, xshift = -0.25 cm]{$F$};
	\node[draw = none, fill = none, below = 0.2of x, xshift = -0.7 cm]{$D$};
	\node[draw = none, fill = none, below = 0.2of x, xshift = 0.7 cm]{$E_2$};
	\node[draw = none, fill = none, below = 0.2of Y, xshift = 0.25 cm]{$E_1$};
	\node[draw = none, fill = none, below = 0.2of Y2, xshift = -0.25 cm]{$E_1$};
	\node[draw = none, fill = none, below = 0.2of X2, xshift = -0.25 cm]{$E_2$};
	\node[draw = none, fill = none, below = 0.2of X, xshift = 0.25 cm]{$E_2$};
	
	\node[draw,fill=none, below =0.5of Dxx] (Discrim){$\mathcal L_{d-rec}$};
	\node[draw, fill=none, below =0.5of Discrim] (Min){$\mathcal L_{sim}$};
	\node[draw , fill = none, below = 0.5of Dyy] (Reg){$\mathcal L_{diff}$};

	\drawdotted {Dxx} {Reg};
	\drawdotted {Dyy} {Reg};
	\drawdotted {X} {Discrim};
	\drawdotted {X2} {Discrim};
	\drawdotted {Z}{Min};
	\drawdotted {Z2}{Min};
	
	\edge {X,X2}{Dxx};
	\edge {Y,Y2}{Dyy};
	
\end{tikzpicture}